\documentclass[11pt,a4paper]{article}
\usepackage[hyperref]{acl2021}
\usepackage{times}
\usepackage{latexsym}

\usepackage{microtype}
\usepackage{times,graphicx,multirow,amsmath,subfigure,caption,amssymb,array,booktabs,xcolor}

\aclfinalcopy 
\def\aclpaperid{3320} 

\title{Hierarchical Task Learning from Language Instructions with Unified Transformers and Self-Monitoring}

\author{Yichi Zhang ~~~~~~~~~~  Joyce Y. Chai\\
  Computer Science and Engineering \\
  University of Michigan\\
  Ann Arbor, MI, USA \\
  \texttt{\{zhangyic, chaijy\}@umich.edu}}

\date{}

\newcommand{\modelname}{HiTUT}

\begin{document}
\maketitle

\begin{abstract}
Despite recent progress, learning new tasks through language instructions remains an extremely challenging problem. On the ALFRED benchmark for task learning, the published state-of-the-art system only achieves a task success rate of less than 10\% in an unseen environment, compared to the human performance of over 90\%. 
To address this issue, this paper takes a closer look at task learning. 
In a departure from a widely applied end-to-end architecture, we decomposed task learning into three sub-problems: sub-goal planning, scene navigation, and object manipulation; and developed a model {\bf HiTUT}\footnote{Source code available at \url{https://github.com/594zyc/HiTUT}} (stands for {\bf Hi}erarchical {\bf T}asks via {\bf U}nified {\bf T}ransformers) that addresses each sub-problem in a unified manner to learn a hierarchical task structure.  
On the ALFRED benchmark, HiTUT has achieved the best performance with a remarkably higher generalization ability. In the unseen environment, HiTUT achieves over 160\% performance gain in success rate compared to the previous state of the art. The explicit representation of task structures also enables an in-depth understanding of the nature of the problem and the ability of the agent, which provides insight for future benchmark development and evaluation. 

\end{abstract}


\section{Introduction}
\label{sec:intro}

\begin{figure}[t]
	\centering
	\includegraphics[width=1.0\columnwidth]{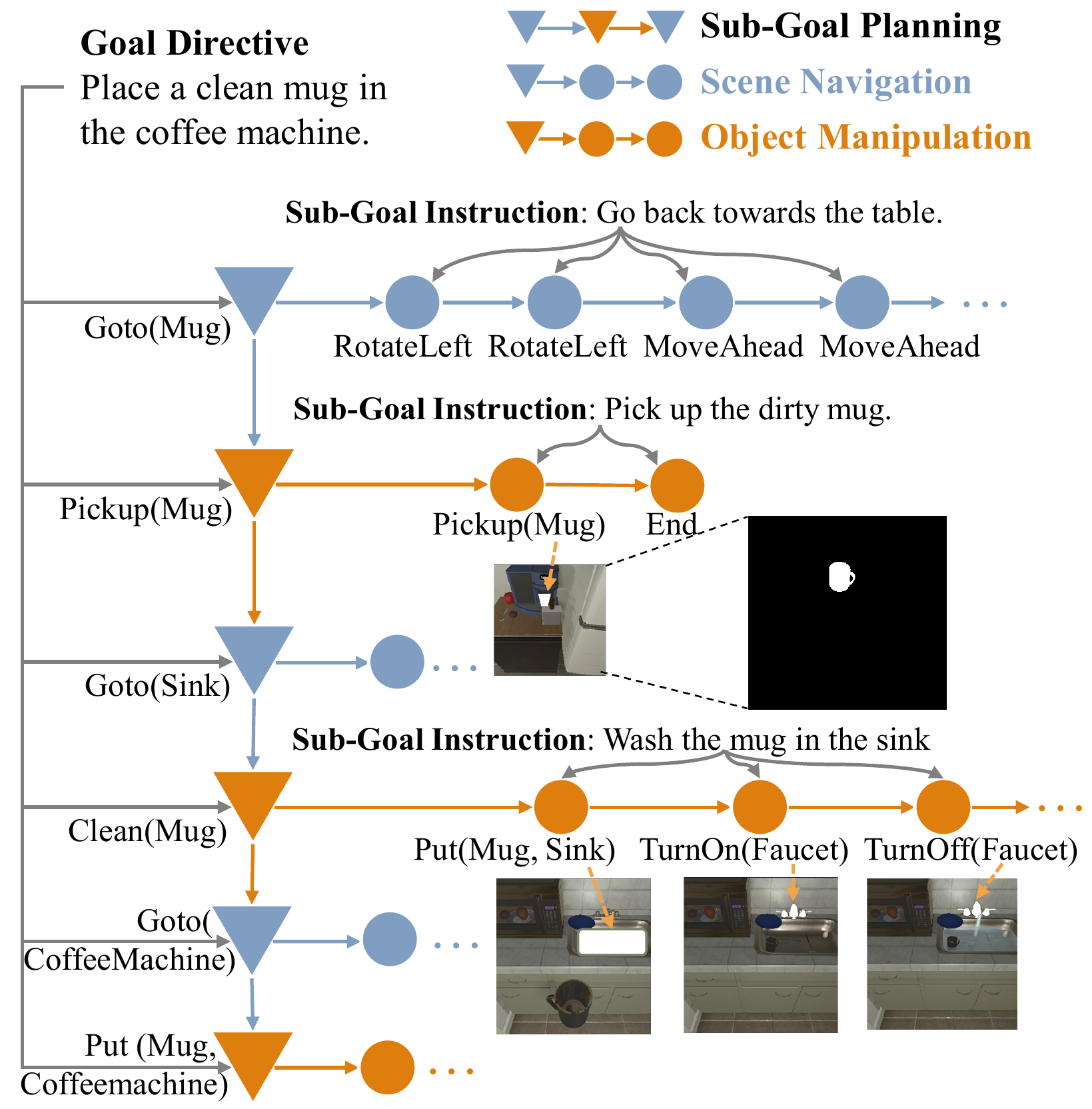}
	\vspace{-2em}
	\caption{\small An example task in ALFRED. }
	\label{fig:subproblems}
	\vspace{-1.5em}
\end{figure}



As physical agents (e.g., robots) start to emerge as our assistants and partners, it has become increasingly important to empower these agents with an ability to learn new tasks by following human language instructions. Many benchmarks have been developed to study the agent's ability to follow natural language instructions in various domains including navigation \cite{anderson2018vln,chen2019touchdown}, object manipulation \cite{misra2017mapping,zhu2017visual} and embodied reasoning \cite{das2018embodied,gordon2018iqa}. Despite recent progress, learning new tasks through language instructions remains an extremely challenging problem as it touches upon almost every aspect of AI from perception, reasoning, to planning and actions. For example, on the ALFRED benchmark for task learning \cite{shridhar2020alfred}, the state-of-the-art system only achieves less than 10\% task success rate in an unseen environment~\cite{pratap2020moca}, compared to the human performance of over 90\%.  Most previous works apply an end-to-end neural architecture \cite{shridhar2020alfred, pratap2020moca,storks2021we} which attempt to map language instructions and visual inputs directly to actions. While striving to top the leader board for end task performance, these models are opaque, making it difficult to understand the nature of the problem and the ability of the agent. 

To address this issue, this paper takes a closer look at task learning using the ALFRED benchmark. 
In a departure from an end-to-end architecture, we have developed an approach to learn the hierarchical structure of task compositions  from language instructions. As shown in Figure~\ref{fig:subproblems}, a high-level goal directive (``place a clean mug in the coffee machine'') can be decomposed to a sequence of sub-goals. Some sub-goals involve navigation in space (e.g., \texttt{\small Goto(Mug)}, \texttt{\small Goto(Sink)}) and others require manipulation of objects (e.g., \texttt{\small Pickup(Mug)}, \texttt{\small Clean(Mug)}). These sub-goals can be further decomposed into navigation actions such as \texttt{\small RotateLeft} and \texttt{\small MoveAhead}, and manipulation actions such as \texttt{\small Put(Mug, Sink)}, \texttt{\small TurnOn(Faucet)}. In fact, such hierarchical structure is similar to Hierarchical Task Network (HTN) widely used in AI planning~\cite{erol1994htn}. While this hierarchical structure is explicit and has several advantages in planning and making models transparent, how to effectively learn such structure remains a key challenge. 

Motivated by recent work in multi-task learning~\cite{liu2019multi}, we decomposed task learning in ALFRED into three sub-problems: sub-goal planning, scene navigation, and object manipulation; and developed a model called {\bf HiTUT} (stands for {\bf Hi}erarchical {\bf T}asks via {\bf U}nified {\bf T}ransformers) that addresses each sub-problem in a unified manner to learn a hierarchical task structure.  On the ALFRED benchmark, HiTUT has achieved the best performance with a remarkably higher generalization ability. In the unseen environment, HiTUT achieves over 160\% performance gain in success rate compared to the previous state of the art. 

The contributions of this work lie in the following two aspects. 

\vspace{3pt}
\noindent
{\bf \em An explainable model achieving the new state-of-the-art performance.} By explicitly modeling a hierarchical structure, our model offers explainability and allows the agent to monitor its own behaviors during task execution (e.g., what sub-goals are completed and what to accomplish next). When a failed attempt occurs, the agent can backtrack to previous sub-goals for alternative plans to execute. This ability of self-monitoring and backtracking offers flexibility to dynamically update sub-goal planning at the inference time to cope with exceptions and new situations. It has led to a significantly higher generalization ability in unseen environments. 


\vspace{3pt}
\noindent
{\bf \em A de-composable platform to support more in-depth evaluation and analysis.} The decomposition of task learning into sub-problems not only makes it easier for an agent to learn, but also provides a tool for an in-depth analysis of task complexity and the agent's ability. For example, one of our observations from the ALFRED benchmark is that the agent's inability to navigate is a major bottleneck in task completion. Navigation actions are harder to learn than sub-goal planning and manipulation actions. For manipulation actions, the agent can learn action types and action arguments predominantly based on sub-goals and the history of actions, while language instructions do not contribute significantly to learning. The success of manipulation actions also largely depends on the agent's ability in detecting and grounding action arguments to corresponding objects in the environment. These findings allow a better understanding of the nature of the tasks in ALFRED and provide insight to address future opportunities and challenges in task learning.

\section{Related Work}

Recent years have seen an increasing amount of work on in the intersection of language, vision and robotics. One line of work particularly focuses on teaching robots new tasks through demonstration and instruction~\cite{rybski07,mohseni2018}. Originated in the robotics community, learning from demonstration (LfD)~\cite{thomaz2009,argall09} enables robots to learn a mapping from world states to robots' manipulations based on human's demonstration of desired robot behaviors. More recent work has also explored the use of natural language and dialogue together with demonstration to teach robots new actions~\cite{mohan2014,scheutz2017, liu2016,she2017, chai2018language, gluck2018interactive}. 


To facilitate task learning from natural language instructions, several benchmarks using simulated physical environment have been made available~\cite{anderson2018vln,misra2018mapping,blukis2019,shridhar2020alfred}. In particular, the vision and language navigation (VLN) benchmark~\cite{anderson2018vln} has received a lot of attention. Many models have been developed, such as the Speaker-Follower model \cite{friedSpeakerfollowerModelsVisionandlanguage2018}, the Self-Monitoring Navigation Agent\cite{ma2019selfmonitoring,ke2019tactical}, the Regretful Agent \cite{ma2019theregretful}, and the environment drop-out model~\cite{tan-etal-2019-learning}. The VLN benchmark is further extended to study the fidelity of instruction following \cite{jainStayPathInstruction2019} and examined to understand the bias of the benchmark~\cite{zhang2020diagnosingvln}. 
Beyond navigation, there are also benchmarks that additionally incorporate object manipulation to broaden research on vision and language reasoning, such as embodied question answering \cite{das2018embodied,gordon2018iqa}. The work closest to ours is the Neural Modular Control (NMC) \cite{das2018neural}, which also decomposes high-level tasks into sub-tasks and addresses each 
 sub-task accordingly. However, self-monitoring and backtracking between sub-tasks is not explored in NMC.


The ALFRED benchmark consists of  high-level goal directives such as \textit{``place a clean mug in the coffee machine''} and low level language instructions such as \textit{``rinse the mug in the sink''} and \textit{``turn right and walk to the coffee machine''} to accomplish these goals. In addition to language instructions, it also comes with expert demonstrations of task execution in an interactive visual environment. We choose this dataset because its unique challenges are closer to the real world, which require the agent to not only learn to ground language to visual perception but also learn to plan for and execute actions for both navigation and object manipulation. 




\section{Hierarchical Tasks via Unified Transformers}
\label{sec:hitut}

As discussed in Section~\ref{sec:intro}, task structures are inherently hierarchical, which compose of goals and sub-goals. Different sub-goals involve tasks of different nature. For example, navigation focuses on path planning and movement trajectories, while manipulation concerns more about interactions with concrete objects. 
Instead of end-to-end mapping from language instructions to primitive actions \cite{shridhar2020alfred,pratap2020moca,storks2021we}, we decomposed task learning into three separate but connected sub-problems: sub-goal planning, scene navigation, and object manipulation, and developed a model called {\bf HiTUT} (stands for {\bf Hi}erarchical {\bf T}asks via {\bf U}nified {\bf T}ransformers) to tie these sub-problems together to form a hierarchical task structure.

\subsection{Task Decomposition}
We first introduce some notations to describe the task and the model. There are three types of information:

\vspace{2pt}
\noindent
- Language ($\mathcal{L}$). We use $G$ to denote a high-level goal directive, e.g., ``place a clean mug in the coffee machine'' and  $I_i$ to refer to a specific low-level language instruction. 

\vspace{2pt}
\noindent
- Vision ($\mathcal{V}$). It captures the visual representation of the environment. 

\noindent
- Predicates ($\mathcal{P}$). Symbolic representations are defined to  capture three types of predicates: sub-goals ($sg$), navigation actions ($a^n$), and manipulation actions ($a^m$). Each $sg$ has two parts ($sg^{type}$, $sg^{arg}$) where $sg^{type}$ is the {\em type} (e.g., \texttt{\small Goto}) and $sg^{arg}$ is the {\em argument} (e.g., \texttt{\small Knife}). Each $a^n$ specifies a type ($a^{n\_type}$) of action, from \texttt{\{\small RotateLeft, RotateRight, MoveAhead, LookUp, LookDown\}}.  Each $a^m$ has also two parts ($a^{m\_type}$, $a^{m\_arg}$) where $a^{m\_type}$ is the action type (e.g., \texttt{\small TurnOn}); $a^{m\_arg}$ is the action argument (e.g., \texttt{\small Faucet}). 

\paragraph{Sub-Goal Planning.}
Sub-goal planning acquires a sequence of sub-goals $sg_1,\cdots,sg_n$ to accomplish the high-level goal $G$. 
We predict the type $sg_i^{type}$ and argument $sg_i^{arg}$ separately to avoid the combinatorial expansion of the output space. 
Previous work~\cite{jansen2020visually} models sub-goal planning merely from high-level goal directives without visual grounding. These plans are fixed and thus not robust to potential failures during execution and variations of the visual environment. 
To overcome these drawbacks, our sub-goal planning is done on the fly after the previous sub-goal is executed in the environment. More specifically, our sub-goal planning objective is to learn a model ($M_{sg}$) that takes the visual observation at the current step ($v_t$), the high-level goal directive ($G$), and a complete sub-goal history prior to the current step ($sg_{<i}$) to predict the current sub-goal as follows: 
$$sg_i\triangleq(sg_i^{type}, sg_i^{arg})=M_{sg}(v_t, G, sg_{<i})$$
The predicted sub-goals serve as a bridge between the high-level goal and the low-level predictions of navigation actions and/or manipulation actions.


\paragraph{Scene Navigation. }
Navigation sub-goals only require predictions for the types of navigation actions. The objective is to learn a model for navigation ($M_{n}$) which takes the current visual observation ($v_t$), current sub-goal ($sg_i$), language instruction ($ I_i$), and the navigation action history up to the current step ($a^{n}_{<j}$) to predict the next navigation action:   
$$a_j^{n} \triangleq a^{n\_type}_j = M_{n}(v_t, I_i, sg_i, a^{n}_{<j}) $$



\paragraph{Object Manipulation. }
For a manipulation sub-goal, in addition to the type and argument of the action, the model ($M_{m}$) also needs to 
generate a segmentation mask ($m_j$) on the current visual observation to indicate which object to interact with (i.e., which object the argument is grounded to):
\begin{align*}
(a_j^{m}, m_j) 
&\triangleq (a^{m\_type}_j, a^{m\_arg}_j, m_j) \\
&= M_{m}(v_t, I_i, sg_i, a^{m}_{<j}) 
\end{align*}


\noindent
The mask prediction is crucial because the action will not be successfully executed with an incorrect grounding even if $a_j^{m}$ is correctly predicted. 

\vspace{3pt}
\noindent
As described above, although the context of the three sub-problems varies, each model has similar input components from the space of  $\langle\mathcal{V}, \mathcal{L}, \mathcal{P}\rangle$. This similarity inspires us to design an unified model to solve three sub-problems simultaneously. 

\subsection{Unified Transformers} 
\label{sec:model}

\begin{figure}[t]
	\centering
	\includegraphics[width=1.0\columnwidth]{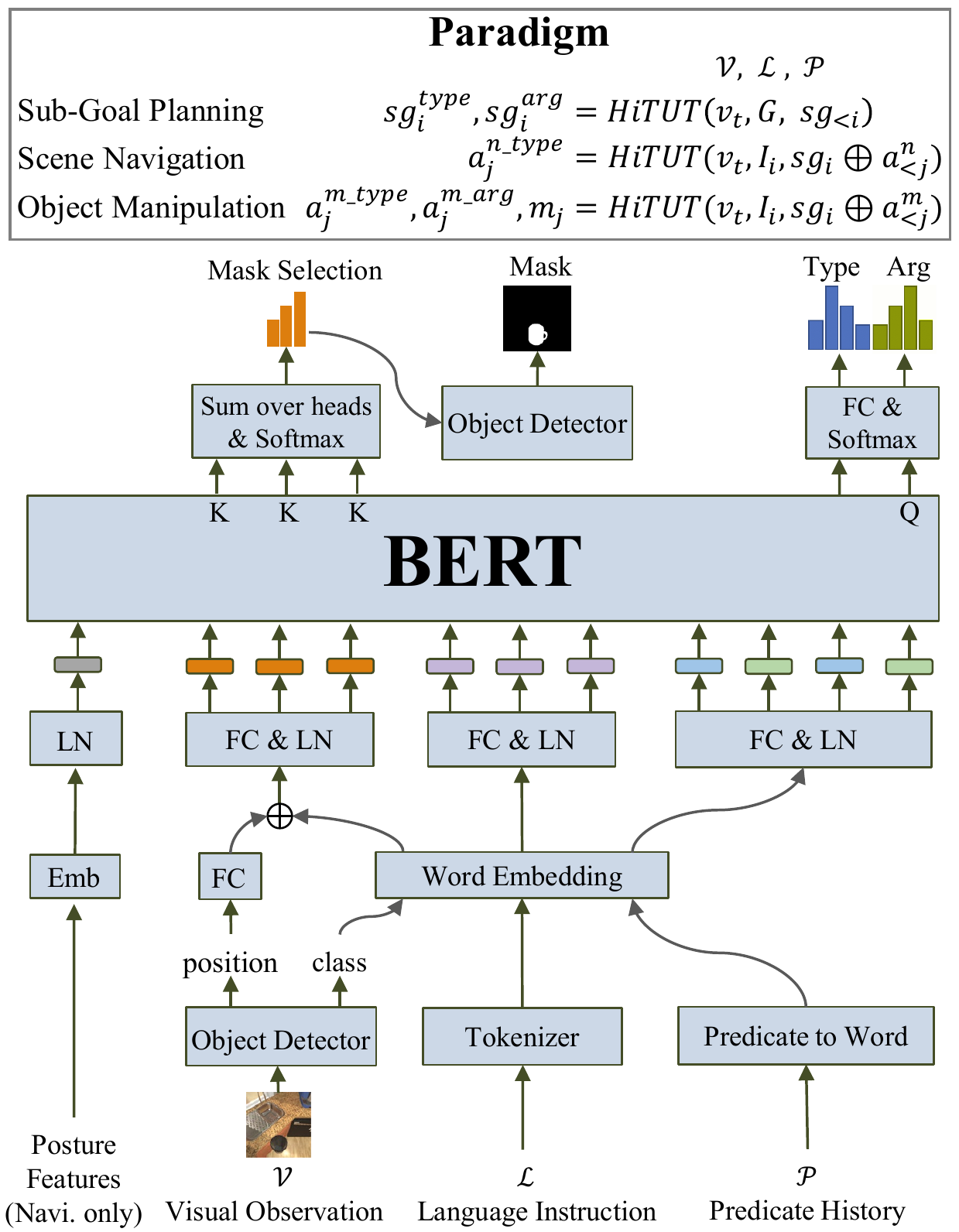}
	\vspace{-1em}
	\caption{\small The structure of HiTUT. }
	\label{model}
	\vspace{-1em}
\end{figure}




We leverage the effective self-attention based model \cite{vaswani2017attention} to capture the correspondence of different input sources as shown in Figure \ref{model}. We first project the input from different modalities into the language embedding space, and adopt a transformer to integrate the information together. Multiple prediction heads are constructed on top of the transformer encoder to make predictions for the sub-goal type and argument, the action type and argument, and object masks respectively. As the three sub-problems share the similar input form, we solve them all together using a unified model based on multi-task learning \cite{liu2019multi}. 

Our model differs from previous works \cite{shridhar2020alfred,pratap2020moca} in the following aspects. First, we do not apply recurrent state transitions, but feed the prediction history as the input to each subsequent prediction. This may help better capture correlations between predicates and other modalities. Second, we do not use dense visual features from the scene, but rather the object detection results. By doing this, we map different modalities to the word embedding space before feeding them into the transformer encoder, thus taking advantage of the pre-trained language models. 
Third, we use a predicate embedding to share linguistic knowledge between predicate symbols and word embeddings. 

\vspace{-5pt}
\paragraph{Predicate Embedding.}
We use the term predicates to refer to symbolic representations including sub-goal types, action types, and their arguments. We map symbols to their corresponding natural language phrases (e.g., \texttt{\small AppleSliced} is mapped to \textit{a sliced apple}). We then tokenize and embed the tokens using word embeddings, and take the sum of the embeddings to obtain the representation of each predicate. 

\vspace{-5pt}
\paragraph{Vision Encoding.}
We use a pre-trained object detector (Mask R-CNN \cite{he2017mask}) to encode visual information. Instead of dense features, we simply use the detection results (class labels, bounding box coordinates and confidence scores) as visual features. Specifically, we use the top $K$ detected objects with a confidence score higher than 0.4 to form the visual features. The object class labels share the same space with object arguments, thus can be embedded into the same space. The position information of an object is encoded by a 7-dimensional vector consisting of its coordinates, width and height of the bounding box and its confidential score. This vector is first mapped to the same dimension as word embeddings by a liner transformation, then added to the class embedding to form the final object representation. 

\vspace{-5pt}
\paragraph{Object Grounding.}
\modelname{} does not generate masks by itself. Instead it chooses an object from the $K$ input objects and uses the corresponding mask generated by the object detector. This method makes use of the strong prior learned from object detection pre-training, so the model can focus on learning the grounding task. 
A drawback is that the object detector cannot be improved during training, and the performance of the detector determines the upper bound of our model's grounding ability. 
We leave the exploration of more robust grounding method for future work.

\paragraph{Posture Feature}
We use an additional posture feature to assist scene navigation, which includes the agent's rotation (N, S, E, W) and its angle of sight horizon (discretized by 15 degree). The positions are embedded and summed up to form the posture feature representation. The agent maintains its own posture in the form of a relative change to its initial posture instead of the absolute posture in the environment, thus avoid using additional sensory data.



\begin{figure}[t]
	\centering
	\includegraphics[width=1.0\columnwidth]{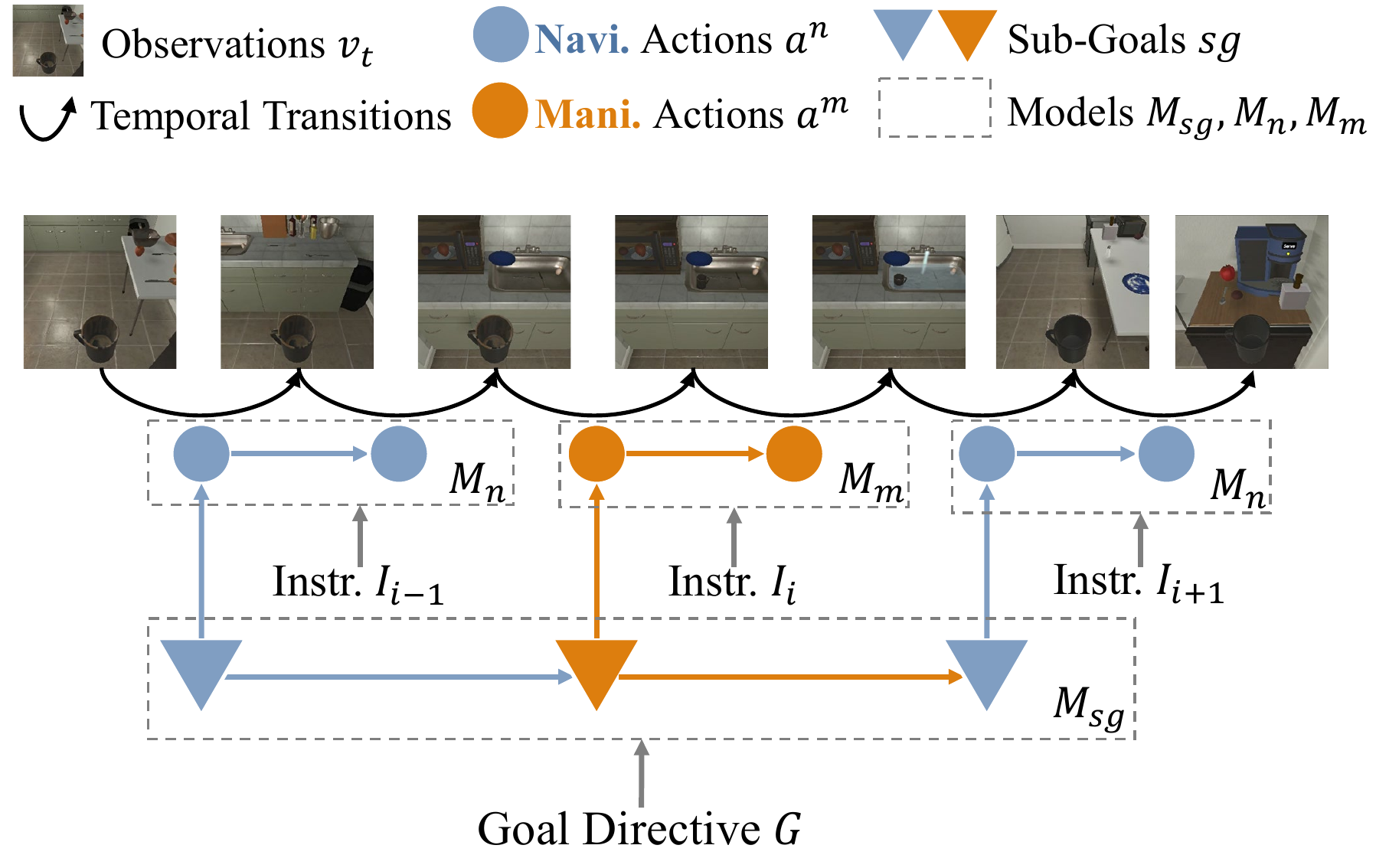}
	\vspace{-2em}
	\caption{\small Overview of HiTUT where unified transformers for sub-programs are integrated together . }
	\label{fig:integration}
\end{figure}

\subsection{Self-Monitoring and Backtracking}
\label{sec:backtracking}

These unified transformers trained for sub-problems are integrated together as shown in Figure~\ref{fig:integration}. One important advantage of intermediate sub-goal representations is to facilitate self-monitoring and backtracking which allows the agent to dynamically adjust the plan to cope with failures during execution. As shown in Section~\ref{sec:experiments}, this feature brings out the most remarkable performance gain compared to the state of the art.

\begin{figure*}[t]
	\centering
	\includegraphics[width=1.99\columnwidth]{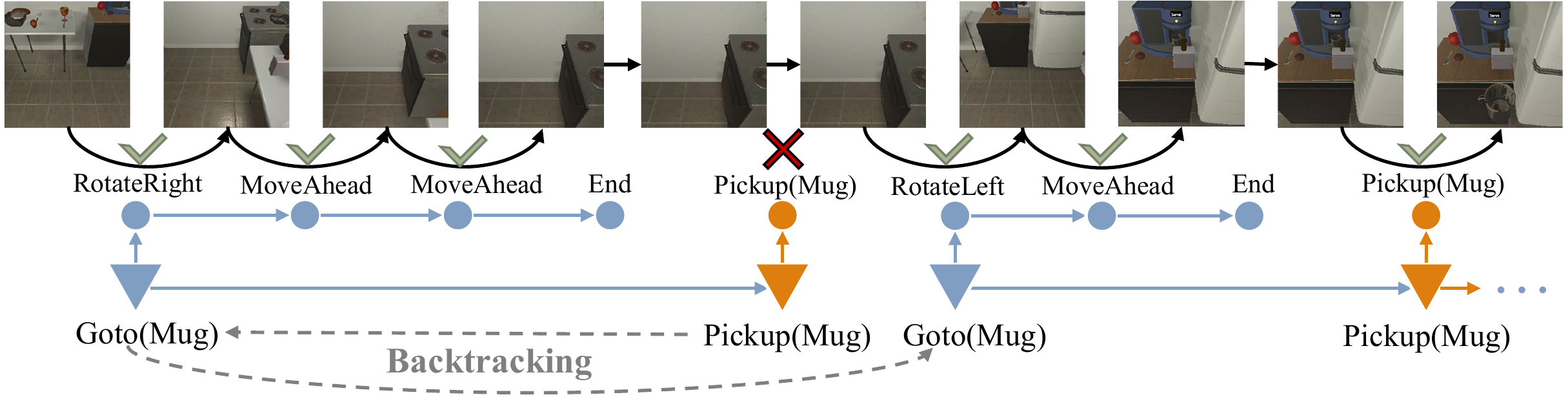}
	\caption{\small Illustration of self-monitoring and backtracking.}
	\label{fig:backtracking}
\end{figure*}

\paragraph{Self-Monitoring.} The world is full of uncertainties, and mistakes are inevitable. Based on the learned model, the agent should be able to monitor its own behaviors and dynamically update its plan when the situation arises. Our explicit representation of sub-goals allows the agent to self-check whether some sub-goals are accomplished. Particularly for manipulation sub-goals, it is feasible for the agent to detect their failures by simply monitoring whether all the manipulation actions are successfully executed. For example, \texttt{\small Clean(Mug)} cannot succeed if any of the actions along the path \texttt{\small Put(Mug, Sink), TurnOn(Facuet), TurnOff(Facuet), Pickup(Mug)} fail. When the agent detects the failure of a subgoal, for example, as shown in Figure~\ref{fig:backtracking} the manipulation sub-goal \texttt{\small Pickup(Mug)} fails, it can reason about whether the previous sub-goal (i.e., \texttt{\small Goto(Mug)}) is successfully achieved. 

\label{sec: backtracking}
\paragraph{Backtracking.} In classical AI, backtracking is the technique to go back and try an alternative path that can potentially lead to the goal. As shown in Figure~\ref{fig:backtracking}, when \texttt{\small Pickup(Mug)} fails, the agent backtracks to  \texttt{\small Goto(Mug)} and tries a different sequence of primitive actions to accomplish this sub-goal. In ALFRED, only based on the visual information without other sensory information (e.g., only observing a mug without knowing how far it is), is it difficult to check whether a navigation sub-goal is successfully achieved (e.g. whether a \texttt{\small Mug} is reachable).
So every time after trying a different path for \texttt{\small Goto(Mug)}, the agent will check whether the subsequent manipulation action \texttt{\small Pickup(Mug)} is successful. If it's successful, the agent will move on to the next sub-goal; otherwise the agent will continue to backtrack until a limit on the maximum number of attempts is reached. Our explicit representation of sub-goals makes this backtracking possible and has led to a significant performance gain in unseen environments.

\begin{table}[t]
\resizebox{.99\linewidth}{!}{
\begin{tabular}{lccccc}
\toprule
           & \multicolumn{1}{c}{Train} & \multicolumn{2}{c}{Validation} & \multicolumn{2}{c}{Test}       \\
           \cmidrule(lr){3-4} \cmidrule(lr){5-6}
           &&Seen & Unseen &Seen & Unseen \\
           \midrule
\#Scenes     & 108 & 88 & 4 & 107 & 8  \\ 
\#Demonstrations  & 6,574 & 251 & 255 & 483 & 488  \\ 
\#Annotations     & 21,023 & 820 & 821 & 1,533 & 1,529  \\ 
\#Sub-goals & 162k & 6.4k & 6.0k & - & - \\
\#Navi. Actions  & 983k & 39k & 35k & - & -\\
\#Mani. Actions   & 209k & 8.3k & 8.1k & - & -\\ \bottomrule
\end{tabular}}
\caption{Statistics of data distribution in ALFRED. The number of annotations is equivalent to the number of tasks in each split.}
\label{tb_dataset} 
\end{table}

\begin{table*}[t]
\centering
\resizebox{.99\linewidth}{!}{
\begin{tabular}{lcccccccc}
\toprule
\multirow{3}{*}{Model}&\multicolumn{2}{c}{Validation Seen} & \multicolumn{2}{c}{Validation Unseen}
&\multicolumn{2}{c}{Test Seen} & \multicolumn{2}{c}{Test Unseen}   \\ 
\cmidrule(lr){2-3} \cmidrule(lr){4-5} \cmidrule(lr){6-7} \cmidrule(lr){8-9} 
& Success & Goal-Cond & Success &Goal-Cond & Success & Goal-Cond & Success &Goal-Cond \\
\midrule
Seq2Seq  & 3.70 (2.10) & 10.00 (7.00) & 0.00 (0.00) & 6.90 (5.10) 
                & 3.98 (2.02) & 9.42 (6.27)  & 0.39 (0.80) & 7.03 (4.26)  \\
HAM         & - & - & - & - 
                & 12.40 (8.20) & 20.68 (18.79)  & 4.50 (2.24) & 12.34 (9.44)  \\
MOCA           & 19.15 (\textbf{13.60}) & 28.50 (\textbf{22.30}) & 3.78 (2.00)  & 13.40 (8.30) 
                & \textbf{22.05} (\textbf{15.10}) & 28.29 (\textbf{22.05}) & 5.30 (2.72) & 14.28 (9.99) \\
\modelname{} & \textbf{25.24} (12.20) & \textbf{34.85} (18.52) & \textbf{12.44} (\textbf{6.85})  & \textbf{23.71} (\textbf{11.98}) 
                        &21.27 (11.10) & \textbf{29.97} (17.41) & \textbf{13.87} (\textbf{5.86})  & \textbf{20.31} (\textbf{11.51}) \\
\midrule
\modelname{} ($G$ only)     &18.41 (7.59) & 25.27 (12.55) & 10.23(4.54) & 20.71 (9.56)
                            & 13.63 (5.57)  & 21.11 (11.00) & 11.12 (4.50) & 17.89 (9.77) \\ 
\midrule
Human &-&-&-&-&-&-&91.00 (85.80) & 94.50 (87.60)\\
\bottomrule
\end{tabular}}
\caption{\small Task and Goal-Condition success rates. The path length weighted version is in parentheses. The highest values per column are in {\bf bold}. "-" denotes scores that are not reported. \textit{G only} denotes only using the goal directive during evaluation without any sub-goal instructions.  
}
\label{tb_benchmark} 
\end{table*}

\section{Experiments}
\label{sec:experiments}

\subsection{Setting and Implementation}
\label{sec: exp_setting}
\paragraph{Dataset.} We follow the train/validation/tests data partition proposed in ALFRED, where validation and test sets are further split into {\em seen} and {\em unseen} based on whether the scene is shown to the model during training. Each sub-goal planning step or a primitive prediction step forms a data instance for the corresponding sub-problem. The number of data instances are shown in Table \ref{tb_dataset}.

\paragraph{Pre-training.} We employ the pre-training followed by fine-tuning paradigm for both the object detector and the main model. For the object detector, we use a Mask R-CNN \cite{he2017mask} model pre-trained on MSCOCO \cite{lin2014microsoft}, and fine-tune it on 50K images collected by replaying the expert trajectories in the ALFRED train split. As we observe that the model struggles on detecting small objects together with large receptacles, we train two networks to detect movable objects and big receptacles separately. We use the pre-trained RoBERTa \cite{liu2019roberta} model to initialize the transformer encoder.

\paragraph{Training. } We perform imitation learning (supervised learning) on the expert demonstrations. The ground-truth labels of sub-goals and primitive actions are obtained from the metadata.  Different input and output labels are organized for each sub-problem respectively as described in Section \ref{sec:hitut}. 
We use the mask proposal that overlaps the most with the ground truth mask as the mask selection label if the intersection-of-union is above 50\%. If there is no valid mask proposals, the label is assigned to 0 as an indicator of non-valid grounding. We optimize the cross-entropy loss between model predictions and the ground truth.
We follow the multi-task training schema in \citet{liu2019multi} where for each iteration, a batch is randomly sampled among all the sub-problems, and the model is updated according to the corresponding objective. More details are in Appendix. 

\vspace{-5pt}
\paragraph{Evaluation Metrics. }
ALFRED leverages an interactive evaluation in the AI2-THOR environment \cite{kolve2017ai2thor}. 
A task is considered successful if all the goal conditions (e.g. the target object is placed on a correct receptacle and in a requested state such as heated or cleaned etc.) are met. Three measures are used: (1) success rate (the ratio of successfully completed tasks), (2) goal-condition rate (ratio of completed goal conditions), and (3) a weight version of these two rates which takes into account of the length difference between the predicted action sequence and the expert demonstrated action sequence \cite{shridhar2020alfred}. 

\vspace{-5pt}
\paragraph{Baselines.}
We compare HiTUT to: (1) Seq2Seq - an LSTM-based baseline model with progress monitoring proposed in \citet{shridhar2020alfred}; (2) HAM - a hierarchical attention model over enriched visual inputs \cite{ham2020eccv}, and (3) MOCA - a modular approach which also uses a Mask R-CNN for mask generation \cite{pratap2020moca} and achieved previous state-of-the-art performance.

\subsection{Evaluation Results}

\subsubsection{Overall Performance of HiTUT}
\label{sec:overall_performance}

We first evaluate the overall performance of the proposed framework as shown in Table \ref{tb_benchmark}. On the testing data reported by the leader board, in seen environments, HiTUT achieves comparable performance as MOCA. However in unseen environments, HiTUT outperforms MOCA by over 160\% on success rate. 
This demonstrates our hierarchical task modeling approach has higher generalization ability compared to end-to-end models. Self-monitoring and backtracking enabled by hierarchical task structures allows the agent to better handle new situations. 
Remarkably, only based on high-level goal directives (i.e., {\em \small HiTUT (G Only)}) without using any sub-goal instructions, is HiTUT able to obtain a success rate of 11\% in unseen environment, achieving 110\% performance gain compared to MOCA. This result indicates that HiTUT can learn prior task knowledge from the hierarchical modeling process and apply that directly in new environment with some success. Nevertheless, our results are far from human performance and there is still huge room for future improvement. 

To have a better understanding of the problem, we also conduct evaluations on sub-goals. The agent is positioned at the starting point of each sub-goal by following the expert demonstration and the success rate of accomplishing the sub-goal is measured.   
HiTUT predicts first a symbolic sub-goal representation and then the action sequence to complete the sub-goal. As shown in Table \ref{tb_subgoal}, HiTUT outperforms previous models on almost all of the manipulation sub-goals by a large margin. The performance gain is particularly significant in unseen environment, which demonstrates the advantage of our explicit hierarchical task modeling in low-level action planning. 

\begin{table}[t]
\resizebox{.99\linewidth}{!}{
\begin{tabular}{lccccccccc}
\toprule
&\multicolumn{1}{c}{Model}  & \rotatebox{70}{Pick} & \rotatebox{70}{Put} & \rotatebox{70}{Cool} & \rotatebox{70}{Heat} & \rotatebox{70}{Clean} & \rotatebox{70}{Slice} & \rotatebox{70}{Toggle} & {Avg.} \\
\midrule
\multirow{3}{*}{\rotatebox{90}{Seen}}
&Seq2Seq  & 32 & \textbf{81} & 88 & 85 & 81 & 25 & \textbf{100}& 70\\
&MOCA     & 53 & 62 & 87 & 84 & 79 & 51 & 93 & 73 \\
&\modelname{}     & \textbf{81} & 77 & \textbf{95} & \textbf{100} & \textbf{83} & \textbf{81} & 97 & \textbf{88}\\
\midrule
\multirow{3}{*}{\rotatebox{90}{Unseen}}
&Seq2Seq & 21 & 46 & 92 & 89 & 57 & 12 & 32 & 50 \\
&MOCA    & 44 & 39 & 38 & 86 & 71 & 55 & 11 & 49 \\
&\modelname{}    & \textbf{71} & \textbf{69} & \textbf{100}& \textbf{97} & \textbf{91} & \textbf{78} & \textbf{58} & \textbf{81} \\
\bottomrule
\end{tabular}}
\caption{\small Success rates of manipulation sub-goals on validation sets. The highest values per fold are in {\bf bold}.}
\vspace{-5pt}
\label{tb_subgoal} 
\end{table}

\subsubsection{The Role of Backtracking}

We conduct experiments to better understand the role of self-monitoring and backtracking. 
We repeat the task-solving evaluation with different limits on the allowed maximum number of backtracking. 
The agent only stops when the model predicts to stop (i.e., predicts \texttt{End}) or it reaches the backtracking limit. 
As shown in Table \ref{tb_backtracking}, as the limit increases, the task/goal-condition success rate increases accordingly. 
One thing notable is that the gap between success rates (weighted and unweighted) 
become larger when more backtrack attempts are allowed. This is within our expectation because backtracking deviates from instruction following navigation to goal-oriented exploration, which usually takes  more steps than the expert demonstration. 

Since backtracking is particularly targeted to navigation sub-goals \texttt{\small Goto} (see Section~\ref{sec: backtracking}), we further examine the role of number of re-tries (i.e. backtracks) in completing the sub-goal. As shown in Table \ref{tb_goto}, HiTUT reaches more targets when given more opportunities to backtrack. The backtracking is most beneficial in unseen environment.


\begin{table}[t]
\centering
\resizebox{.99\linewidth}{!}{
\begin{tabular}{ccccc}
\toprule
&\multicolumn{2}{c}{Valid Seen} & \multicolumn{2}{c}{Valid Unseen} \\
\cmidrule(lr){2-3} \cmidrule(lr){4-5} 
\#BT& Success & Goal-Cond & Success &Goal-Cond \\
\midrule

No & 10.5 (6.0) & 18.4 (13.8) & 5.2 (3.0)  & 13.5 (11.1) \\
2  & 18.9 (9.9) & 27.6 (18.0) & 10.2 (5.9) & 20.2 (13.6)  \\
4  & 23.1 (11.3) & 32.9 (18.6) & 12.9 (7.0)  & 22.7 (12.9) \\
6  & 25.6 (12.0) & 35.1 (18.5) & 14.5 (7.4)  & 24.3 (12.3) \\
8  & 27.2 (12.5) & 37.0 (18.5) & 16.2 (7.8)  & 25.9 (12.1) \\
\bottomrule
\end{tabular}}
\caption{\small Success rates w.r.t. the allowed maximum backtracking number (\#BT).}
\label{tb_backtracking} 
\end{table}

\begin{table}[t]
\resizebox{.99\linewidth}{!}{
\begin{tabular}{lcccccccc}
\toprule
&\multirow{2}{*}{Seq2Seq}&\multirow{2}{*}{MOCA}&\multicolumn{6}{l}{Our model with different \#backtracks} \\
& &  &no&1&2&4&6&8 \\
\midrule
Seen & 51 & 54 &35 & 48 & 56 & 64 & 68 & 70 \\
Unseen &22 &32 &31 & 45 & 53 & 60 & 63 & 65 \\
\bottomrule
\end{tabular}}
\caption{\small Success rate of the navigation sub-goal \texttt{Goto} with backtracking . }
\label{tb_goto} 
\end{table}


\subsubsection{Complexity of Tasks}

\begin{table}[t]
\centering
\resizebox{.99\linewidth}{!}{
\begin{tabular}{lcccc}
\toprule
\multirow{2}{*}{Method}&\multicolumn{2}{c}{Valid Seen} & \multicolumn{2}{c}{Valid Unseen} \\
\cmidrule(lr){2-3} \cmidrule(lr){4-5} 
& Success & Goal-Cond & Success &Goal-Cond \\
\midrule
\modelname{} & 25.2 (12.2) & 34.8 (18.5) & 12.4 (6.8)  & 23.7 (12.0) \\
\midrule
+ Oracle & & & & \\
SG     & 29.0 (15.6)  & 39.1 (21.3) &14.0 (7.6) &25.6 (12.7) \\
N         & 75.0 (72.7)  & 78.0 (77.4) & 57.9 (60.0)  & 67.7 (65.2) \\
SG+N   & 79.2 (77.8)  & 84.0 (81.3) & 64.2 (64.2)  & 72.0 (68.1) \\
SG+N+M     & 89.0 (100)  & 90.0 (100) & 80.5 (100)  & 83.7 (100) \\
SG+N+GR     & 99.3 (99.0) & 99.4 (99.1) & 99.4 (99.3) & 99.6 (99.6)\\
\bottomrule
\end{tabular}}
\caption{\small Success rates of HiTUT with different parts of predictions replaced by oracle operations with expert demonstrations. N, M, SG and GR denote oracle navigation actions, manipulation actions, sub-goals and object grounding (i.e., mask generation) respectively.}
\vspace{-5pt}
\label{tb_oracle} 
\vspace{-5pt}
\end{table}

\begin{figure*}[t]
	\centering
		\subfigure[Sub-Goal Type ]
		{	\label{fig_sg_type}
			\includegraphics[width=0.63\columnwidth]{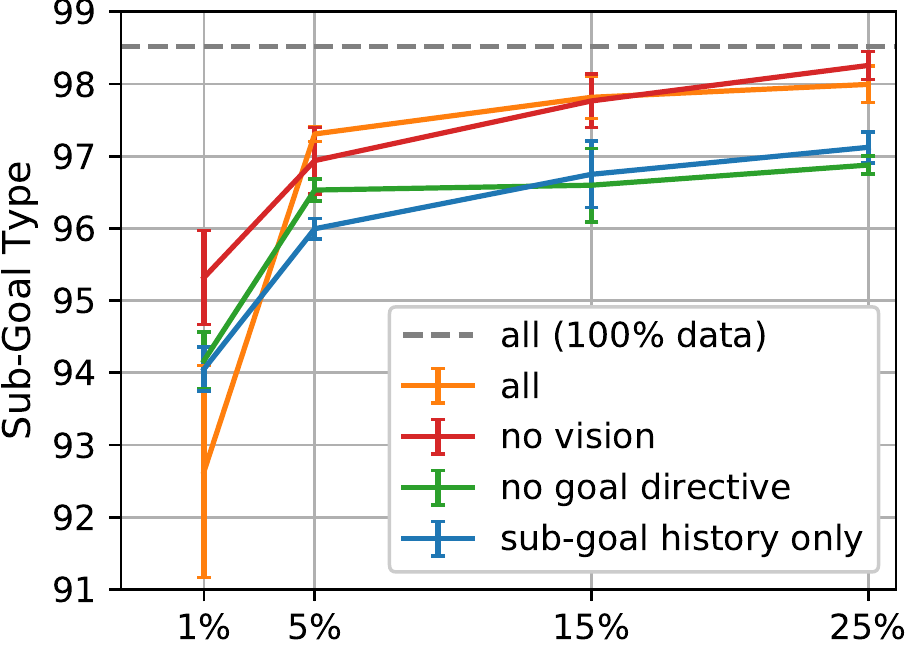} }
		\subfigure[Sub-Goal Argument ]
		{	\label{fig_sg_arg}
			\includegraphics[width=0.63\columnwidth]{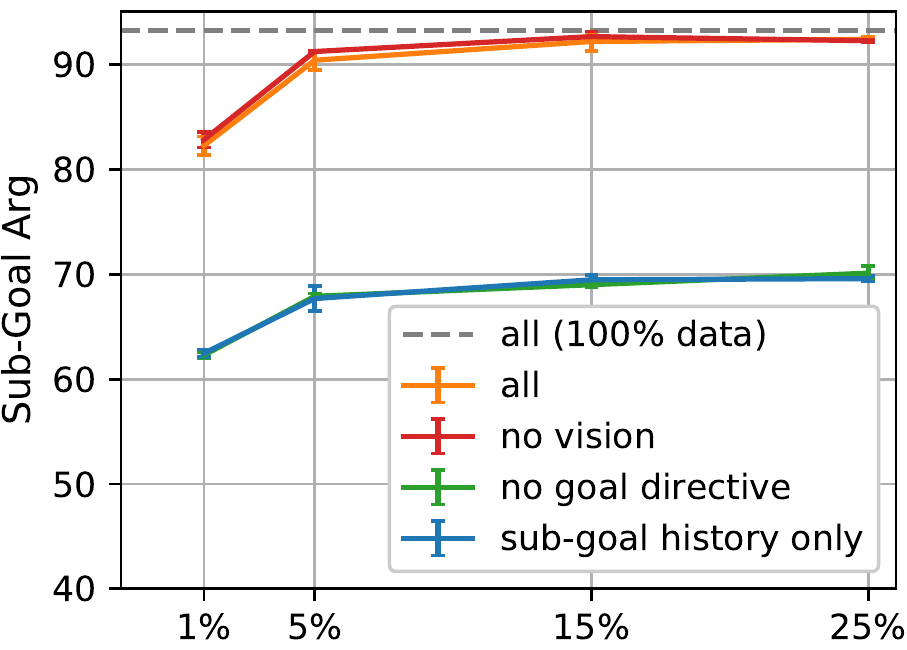} }
		\subfigure[Navigation Action Type]
		{	\label{fig_navi_type}
			\includegraphics[width=0.63\columnwidth]{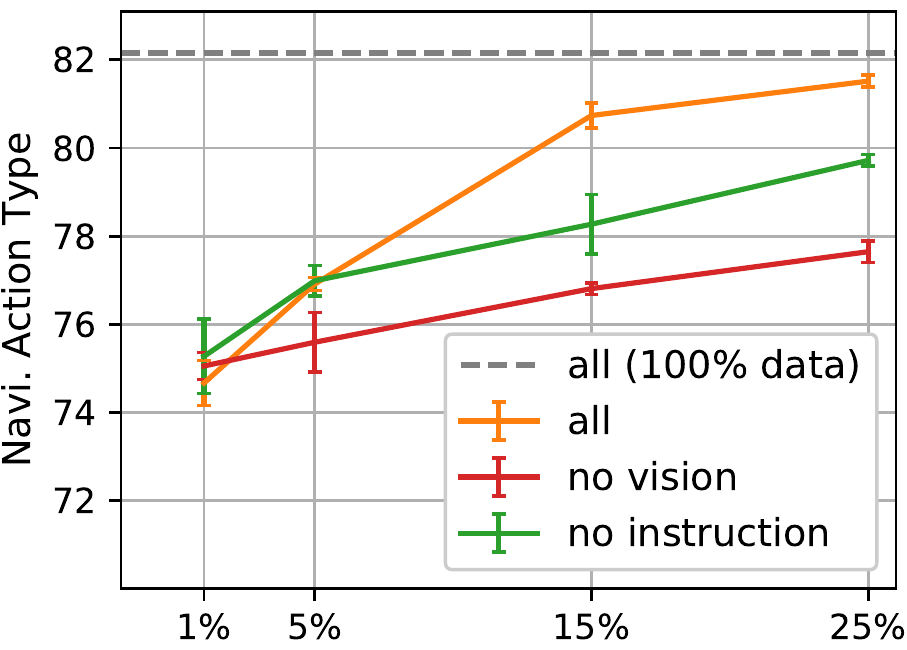} }
		\subfigure[Manipulation Action Type ]
		{	\label{fig_mani_type}
			\includegraphics[width=0.63\columnwidth]{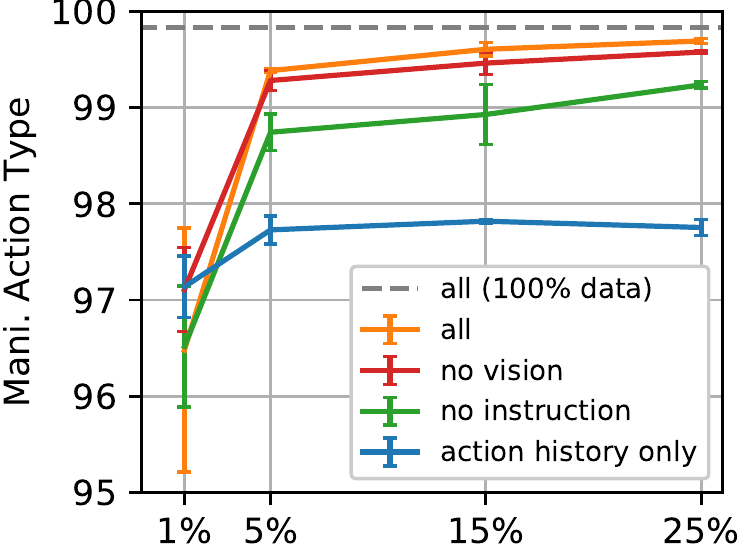} }
		\subfigure[Manipulation Action Argument]
		{	\label{fig_mani_arg}
			\includegraphics[width=0.63\columnwidth]{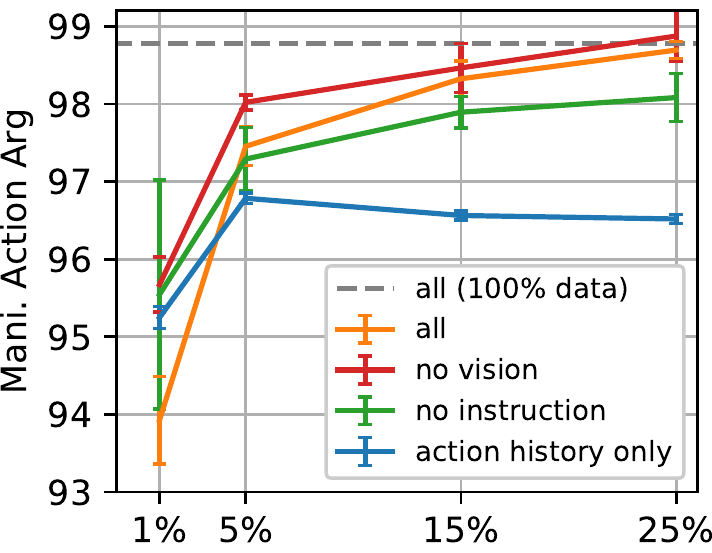} }
		\subfigure[Manipulation Mask Selection ]
		{	\label{fig_mani_mask}
			\includegraphics[width=0.63\columnwidth]{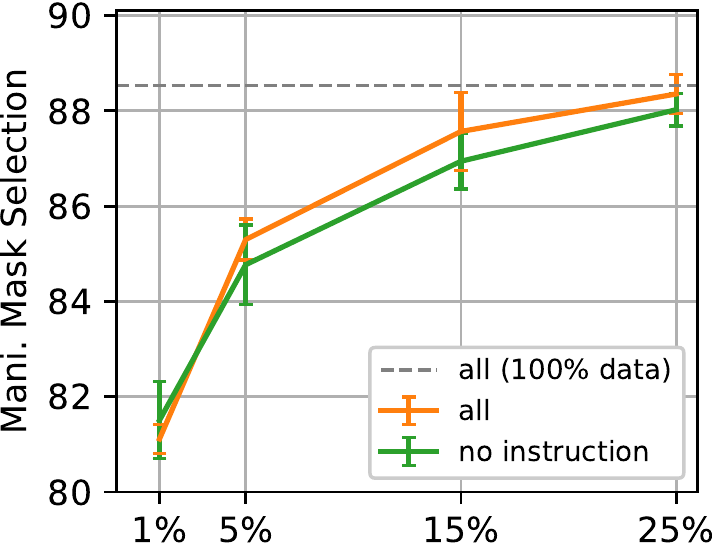} }
	\caption{\small Step-by-step prediction accuracies given the golden sub-goal/action history w.r.t. the proportion of training data on the unseen validation set. Each solid line corresponds to a specific input configuration. Dashed lines are the scores obtained using 100\% of training data.}
	\vspace{-10pt}
	\label{fig_curves}
\end{figure*}

Task decomposition provides a tool to enable better understanding of task complexity and agent's ability. 
To do that, we replace different part of model predictions by the corresponding oracle sub-goals, actions, or masks, as shown in Table \ref{tb_oracle}. 

Using oracle sub-goals improves the success rate for 2\%-6\% (line {\small SG}), showing sub-goal planning is a relatively easy problem and the agent can perform reasonably well. 
After using the oracle navigation actions, the seen and unseen success rates are boosted by an absolute gain of 50\% and 46\% respectively (line {\small N}), indicating that navigating to reach target objects is a particularly hard problem and the agent performs poorly. When oracle sub-goals, navigation actions, and manipulation actions (only symbolic representations) are given (line {\small SG+N+M}), the task success is bounded by the performance of the pre-trained object mask generator (i.e., visual grounding of the object). When oracle object masks are given together with oracle sub-goals and navigation actions (line {\small SG+N+GR}) and the agent only needs to predict symbolic representation of manipulation actions, the performance is near perfect. These last two lines indicate that predicting the type and the argument of a manipulation action is a rather simple problem in the ALFRED benchmark while grounding action arguments to the visual environment remains a challenging task. 


We further examine the complexity of learning to solve sub-problems by evaluating the next-step prediction accuracy given the golden history under different conditions as shown in Figure \ref{fig_curves}. The models are trained and evaluated with different combinations of input and different amount of training data. We observe that excluding the visual input does not hurt performance for sub-goal prediction and manipulation action prediction (shown by a,b,d,e). 
This indicates that in ALFRED, pure symbolic planning is often independent from visual understanding, which is consistent with the findings in \citep{shridhar2020alfworld}. However, this could be an oversimplification brought by the bias in the dataset rather than a true reflection of the physical world. For example, next action prediction can be made by remembering the correlation of predicates instead of reasoning over vision and language, due to the lack of diversity of the task environments.
Removing language instructions causes a minimal performance drop of 1\%-2\% on action prediction tasks, which brings up the question about the usefulness of language instructions in this benchmark. 
Furthermore, the prediction accuracy is above 90\% and 98\% with only 5\% training data for sub-goal and manipulation planning respectively, while the navigation accuracy is only 82\% given all the data. This again supports the finding that planning and performing navigation actions is a much harder problem than sub-goal planning and manipulation actions in ALFRED.  





\section{Discussion and Conclusion}
\vspace{-5pt}

This paper presents a hierarchical task learning approach that achieves the new state-of-the-art performance on the ALFRED benchmark. The task decomposition and explicit representation of sub-goals enable a better understanding of the problem space as well as the current strengths and limitations. Our empirical results and analysis have shown several directions to pursue in the future. 
First, we need to develop more advanced component technologies integral to task learning, e.g., more advanced navigation modules through either more effective structures \cite{hong2020a} or richer perceptions \cite{shen2019situational} to solve navigation bottleneck. 
We need to develop better representations and more robust and adaptive learning algorithms to support self-monitoring and backtracking. We also need to seek ways to improve visual grounding, which is crucial to both navigation and manipulation.  

Second, we should also take a closer look at the construction and objective of existing benchmarks. How a benchmark is created and how truthfully it reflects the complexity of the physical world would impact the scalability and reliability of the approach in the real world. As for the objective,
there is a distinction between {\em learning to perform tasks} and {\em learning to follow language instructions}. If the objective is the former, the agent should be measured by the ability to learn to accomplish high-level goal directives without being given specific language instructions at the inference time. If the objective is the latter, then the agent should be measured by how faithful it follows human instructions aside from achieving the goals, similar to ~\cite{jainStayPathInstruction2019}. 
We need to be clear about the objectives and develop evaluation metrics accordingly. 


Finally, when humans perform poorly in a complex task, we have the ability to diagnose the problem and put more energy on learning the difficult part. Physical agents should also have similar abilities. In task learning, on the one hand, the agent should be able to master simple sub-tasks from a few data instances, e.g., through a few turns of interactions with humans \cite{karamcheti2020learning}. On the other hand, it should be aware of the bottleneck of its learning progress and proactively request for help when problems are encountered either during learning or during deployment \cite{she2017}. How to effectively design interactive and active learning algorithms for the agent to learn complex and compositional tasks remains an important open research question. 









\section*{Acknowledgments}

This work is supported by the National Science
Foundation (IIS-1949634). The authors would like to thank the anonymous reviewers for their valuable comments and suggestions.





\vspace{-15pt}
\bibliographystyle{acl_natbib}
\bibliography{acl2021,itl}

\clearpage

\appendix
\section*{Appendix}

\section{Additional Training Details}
We use the RoBERTa \cite{liu2019roberta} implementation from Huggingface \cite{wolf2019huggingface}. The model is fine-tuned for 10 epochs with the Adam \cite{kingma2014adam} optimizer on the ALFRED training set. The learning rate warms up over the first half of the first epoch to a peak value of 1e-5, and then linearly decayed. The model achieving the highest navigation action prediction accuracy on the validation seen set is selected for evaluation. All the models are trained on one NVIDIA V100 16GB GPU.

\section{Additional Evaluation Details}
We follow the evaluation setting in the ALFRED benchmark\footnote{\url{https://leaderboard.allenai.org/alfred/submissions/get-started}}. For each episode, the agent is given a task, which is composed of a goal directive $G$ and several sub-goal instructions. The agent needs to sequentially perform actions to achieve the goal based on the visual observations of RGB image only. This progress ends if the agent predicts an \textit{End} action (an \textit{End} sub-goal for HiTUT), which is made after up to 10 failed interaction attempts or reaching the maximum step limitation. For HiTUT, there is also a maximum number of backtracking attempts, and the model will be forced to stop if the budget runs out. The maximum number of backtracking is 8 in all of our experiments without explicitly mentioning the backtracking number. We also leverage two techniques to reduce the interaction attempt failures. We use the obstruction detection trick proposed in \citet{pratap2020moca} to avoid failures caused by repeated tries of moving toward obstructions. We propose a self-monitoring approach to check the validity of manipulation actions. If no mask is selected or a predicted action argument is not consistent with the class prediction from Mask R-CNN for the selected object, the manipulation action is decided as failed and the agent performs a backtrack without trying to execute the action. 
Note that in Table \ref{tb_backtracking}, we remove the interaction attempt constraint when comparing the effect of different allowed maximum backtracking numbers, thus the results of $\#BT = 8$ is slightly higher than those shown in Table \ref{tb_benchmark}.

\section{Additional Results}
A detailed per-task performance comparison of HiTUT and MOCA is shown in Table \ref{tb_tasktype}. As the comparison might be unfair since HiTUT benefits from model pre-training, we also conduct an ablation study to show the effectiveness of pre-training. In Table \ref{tb_pretrain}, we compare the fine-tuned RoBERTa model to a Transformer with the same size trained from scratch to show the role of the RoBERTa pretraining.We can see that RoBERTa consistently improves the performance over training from scratch both w/o and w/ backtracking with an absolute gain between 0.4\% and 5\% on task success rate. Notably, Scratch with 4 or 8 backtrackings still outperform MOCA by a large margin in terms of the unseen success rate.

\begin{table}[t]
\resizebox{.99\linewidth}{!}{
\begin{tabular}{lcccc}
\toprule
Task-Type   &  \multicolumn{2}{c}{MOCA} & \multicolumn{2}{c}{HiTUT}       \\
\cmidrule(lr){2-3}\cmidrule(lr){4-5}
               &Seen & Unseen &Seen & Unseen \\
               \midrule
Pick \& Place & 29.5 & 5.0 &\textbf{35.9}&\textbf{26.0}\\
Cool \& Place & \textbf{26.1} & 0.7 &19.0&\textbf{4.6}\\
Stack \& Place& 5.2 & 1.8&\textbf{12.2}&\textbf{7.3}\\
Heat \& Place & \textbf{15.8} & 2.7 &14.0&\textbf{11.9}\\
Clean \& Place &22.3 & 2.4 &\textbf{50.0}&\textbf{21.2}\\
Examine \& Place &20.2 & \textbf{13.2}&\textbf{26.6}&8.1\\
Pick Two \& Place &11.2 & 1.1&\textbf{17.7}&\textbf{12.4}\\
\midrule
Average &18.6 & 3.8 &\textbf{25.2}&\textbf{12.4}\\
\bottomrule
\end{tabular}}
\caption{Success rates across 7 task types on the validation sets. Highest values per fold are {\bf bold}.}
\label{tb_tasktype} 
\end{table}

\begin{table}[t]
\resizebox{.99\linewidth}{!}{
\begin{tabular}{lccc}
\toprule
Model   & \#Backtracking & Seen SR & Unseen SR \\
               \midrule
RoBERTa &      no&                  10.5             & 5.2 \\
Scratch &          no&                   7.9        &       2.8 \\
\hline
RoBERTa &       4     &              23.1          &   12.9 \\
Scratch &           4  &                18.1      &       10.2 \\
\hline
RoBERTa &       8       &            27.2        &     16.2  \\
Scratch &           8    &              26.8    &         14.0  \\
\hline
MOCA     &       -       &             19.15        &    3.78 \\
\bottomrule
\end{tabular}}
\caption{The validation success rates for models pretrained and trained from scratch with different allowed maximum number of backtrackings. }
\label{tb_pretrain} 
\end{table}


\end{document}